\documentclass[showpacs,aps,pre,numerical,twocolumn,10pt,superscriptaddress]{revtex4-1}
\usepackage{graphicx}
\usepackage{amsmath,amssymb}
\usepackage{times,mathptmx}
\usepackage{units}
\usepackage{epsfig}
\usepackage{booktabs}
\usepackage{multirow}

\usepackage[latin1]{inputenc}
\usepackage[english]{babel}
\usepackage{color}

\newcommand{\ii}{\mathrm{i}}

\definecolor{darkgreen}{rgb}{0,0.5,0}
  
\begin{document}

\date{\today}
\title{Reservoir computing with simple oscillators: Virtual and real networks}

\author{Andr\'e Röhm}\affiliation{Institut f{\"u}r Theoretische Physik, Technische Universit{\"a}t Berlin, 10623 Berlin, Germany}
\author{Kathy L{\"u}dge}\affiliation{Institut f{\"u}r Theoretische Physik, Technische Universit{\"a}t Berlin, 10623 Berlin, Germany}

\begin{abstract}
The reservoir computing scheme is a machine learning mechanism which utilizes the naturally occuring computational capabilities of dynamical systems. One important subset of systems that has proven powerful both in experiments and theory are delay-systems. In this work, we investigate the reservoir computing performance of hybrid network-delay systems systematically by evaluating the NARMA10 and the Sante Fe task.. We construct 'multiplexed networks' that can be seen as intermediate steps on the scale from classical networks to the 'virtual networks' of delay systems. We find that the delay approach can be extended to the network case without loss of computational power, enabling the construction of faster reservoir computing substrates.
\end{abstract}


\maketitle

\section{INTRODUCTION}
\label{SEC_intro}

Reservoir computing is a supervised machine-learning scheme for recurrent networks that utilizes the naturally occuring computational power of large dynamical systems. Where more general machine learning schemes aim to train a recurrent neural network in its entirety, reservoir computing differs in its approach by training only a few select links. This divides the system into an input layer, a dynamical reservoir and an output layer. Originally, reservoir computing was inspired both by systematic machine-learning considerations \cite{JAE01} as well as the human brain\cite{MAA02}. It was later found that under certain conditions even a \textit{general} training scheme for recurrent networks can produce structures that mimic the tripartite division of reservoir computing\cite{SCH04g}.

Proposed applications include channel equalizations for satellite communications \cite{BAU15}, real-time audio processing \cite{KEU17} and unscrambling of bits after long-haul optical data transmission \cite{ARG17}.

Reservoir computing works with many different types of dynamical reservoirs. It has also been experimentally demonstrated in a wide variety of systems, benefiting from the fact that the dynamical system need not be trained. Successful demonstrations include systems of dissociated neural cell cultures\cite{DOC09}, a bucket of water \cite{FER03} and field programmable gate arrays (FPGAs)\cite{ANT16}.

Understanding the deeper mechanisms behind the performance of different dynamical reservoirs is still an open problem. Previous works have focused on the link between performance and the Lyapunov Spectrum \cite{ROD11}  and comparison of different node types \cite{VER07}. Extensions of the reservoir computing have also been proposed: Both the use of plasticity \cite{STE07c} of links in the artificial neural network, as well as deterministicly constructing networks \cite{ROD12} try to boost the performance. However, most of these theoretical investigations have focused on the more machine-learning inspired time-discrete artificial neural networks, as opposed to photonic and time-continuous systems. 

Interest in reservoir computing was renewed especially in the photonics and semiconductor community, after Appeltant \textit{et~al.} presented a novel scheme\cite{APP11}. Instead of an extended physical system or large network of single units, this \textit{virtual network} approach uses a long delay-line to produce a high dimensional phase-space in \textit{time}. Several groups have successfully implented such a delay-line based reservoir computer using both optic \cite{BRU13a, VIN15, NGU17} and opto-electronic \cite{LAR12, PAQ12} experiments. Possible extensions to the virtual network scheme have also been considered, among others are hierarchical time-multiplexing \cite{ZHA14a} or the use of counter-propagating fields in a ring laser for simultaneous tasks \cite{NGU15}. Additionally, Ref.~\cite{SCH13l} proposes to analytically calculate the response and time series of such a virtual network and then use the analytic formula to speed up computation.

Simultaneously the traditional network implementation has seen additional improvements by the use of fully passive dynamical reservoirs\cite{VAN08a, VAN11c, VAN14}, which greatly reduces the noise and improves performance. However, for every real node a feed-in mechanism for data, as well as a read-out mechanism for the dynamical response is needed. So even a network of passive elements will still require a significant amount of complexity.
In this paper we aim to show a systematic comparison between the 'delay-line' approach of virtual networks and the original 'real' networks consisting of multiple oscillators. Additionally, we propose a mixed scheme containing both multiple real nodes connected in a network, as well as long delay lines extending the system dimension in time. 

\section{RESERVOIR COMPUTING}
\label{modelrc}

Reservoir computing is a machine learning scheme aiming to utilize the intrinsic computational power of complex dynamical systems. The typical problem is to transform or extract data from a given time-dependent data stream. Usually, the target transformation is not explicitly known or computationally very costly and therefore a direct solution of the problem on a computer seems undesirable. Hence a supervised machine learning approach is used. The learning takes place in the typical two step process: First, a training phase fixes the malleable parameters of the reservoir computer at an optimal value, and then a testing phase evaluates the quality of the learned behaviour.

\begin{figure}
  \begin{center}
    \includegraphics[width=0.45\textwidth]{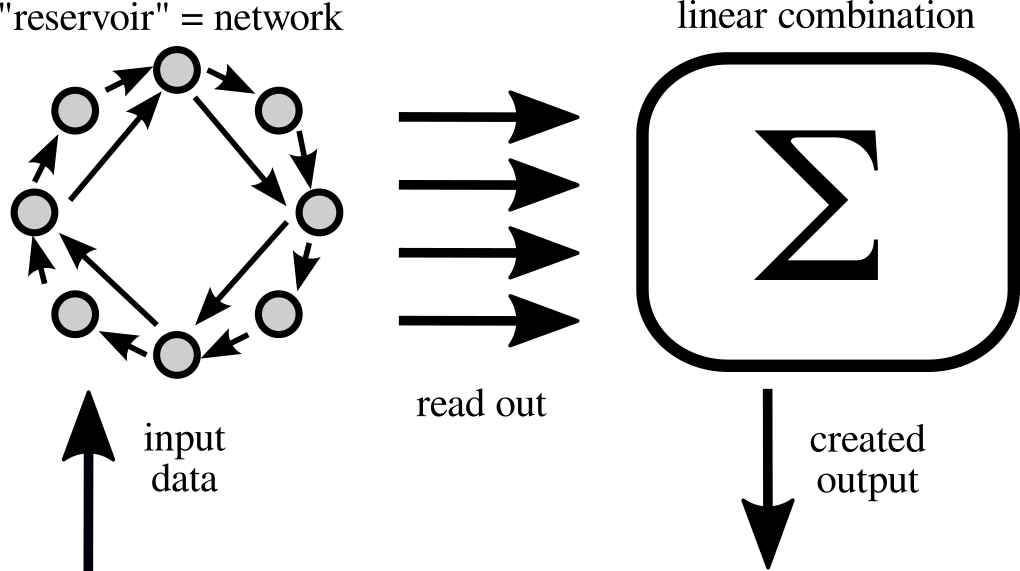}
  \end{center}
\caption{Sketch of the reservoir computing scheme: A stream of input data is fed to a dynamical system, which reacts and traces out a transient in its high-dimensional phase-space. This high-dimensional transient is then recorded and linearly combined to generate the desired output data. }
\label{RC_sketch}
\end{figure}

Figure~\ref{RC_sketch} depicts a sketch of the reservoir computing paradigm. At the core of the reservoir computer lies a dynamical system with a high phase-space, also called 'reservoir'. Historically, these systems were first envisioned to be networks of discrete maps \cite{JAE01} or neural models \cite{MAA02}. The data is fed into the system via some number of parameters, e.g. the driving current of a laser or the voltage applied to neurons, or injected with a driving signal, e.g. input light pulses into an optical system. The dynamical system will then be driven by input data, resulting in some trajectory in its phase space. This process is often called 'expansion in feature space', as the resulting trajectory can be of a much higher dimension than the original data series. The high dimensional response of the dynamical reservoir is then read out and used as the basis for reconstructing the desired output. While conventional deep convolutional neural network learning schemes heavily focus on the training of the internal degrees of the network, the 'reservoir' is assumed to be fixed for reservoir computing. In fact, training is only applied to the linear output weighting, for which a simple linear regression is enough to find the optimal values \cite{JAE01}.

However, for this simplification in the training procedure a price must be paid in system size: We require the desired transformation to be constructable by a mere linear combination of the degrees of freedom of the reservoir. Hence, to be sufficiently computationally powerful, the system needs to be large enough to contain many degrees of freedom. While conventionally trained artificial neural networks are 'condensed' to contain only useful elements, reservoir computing, even in its fully trained state, can carry a lot of overhead, i.e. elements that are not useful for the computation. Therefore reservoir computers can be expected to be always larger than their fully trained counterparts. A simple example of a time-discrete reservoir computer, often called 'echo state network', is shown in the methods section.

However, the fixed nature of the reservoir also allows experimentalists to utilize naturally occuring complex dynamical systems as reservoirs. This is the great advantage of the reservoir computing paradigm. The intrinsic computational powers of physical processes can be used \cite{CRU10a}.

\section{RESERVOIR COMPUTING with delay}
\label{SEC_rc_with_del}

A class of systems that is naturally suited for reservoir computing are delay systems \cite{APP11,  CRU10a}, which are described by delay-differential equations (DDEs). These DDEs contain terms that are not only dependent on the instantaneous variables $X (t) \in \mathbb{R}^N$, but also on their delayed states before a certain time $X (t - \tau)$. Mathematically the phase space dimension of a DDE system is infinite. Many systems in nature can be described by systems of DDEs, where the delay-term usually hides a compressed spatial variable. The most common example are laser systems, where a laser with delayed self-feedback via a mirror has been studied extensively in the literature \cite{LAN80b, ALS96}. Here, the emitted electromagnetic waves can be described with the help of a delay term. Similarly, lasers can also be delay-coupled \cite{SOR13}. Optical and electro-optical systems consisting of only a single node with delay have been successfully used for reservoir computing, and are especially suited due to their high speeds.

\begin{figure}
  \begin{center}
    \includegraphics[width=0.47\textwidth]{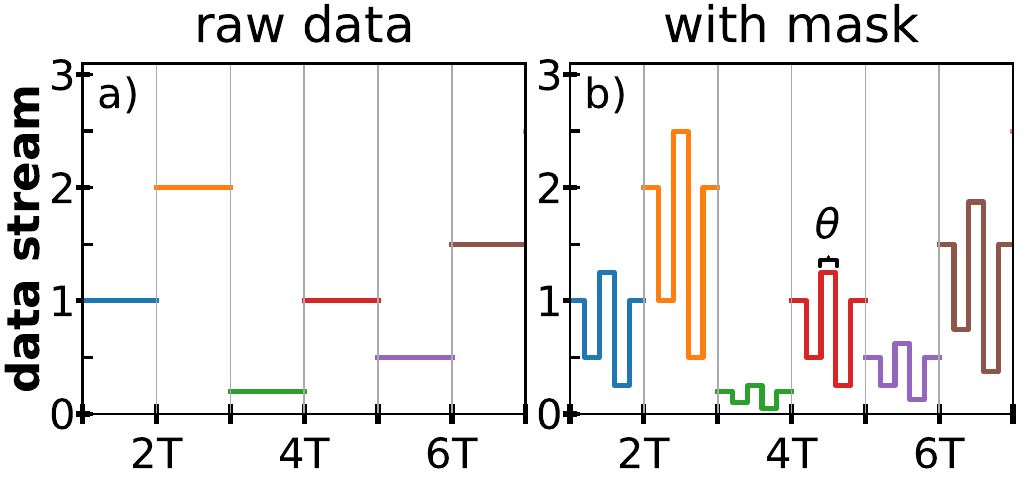}
  \end{center}
\caption{Sketch of the masking or time-multiplexing procedure. The raw input data as given by a vector is transformed into a piece-wise constant function (a). A repeating pattern is multiplied on top of it, called a 'mask'. The resulting masked signal (b) is injected into the system to evoke a more complex phase space response.}
\label{Fig_masking_sketch}
\end{figure}

In reality, measurement resolution and noise limit the amount of information that can be stored within a delay-system. Hence, a single delay-system is not infinitely computationally powerful. (If these limitations apply to the pure mathematical construct of a DDE does not seem obvious). Additionally, real-world systems operate in continuous time - not the discrete time of most of the simulated artifical neural networks. This neccessitates the use of an external clock time and a refined data injection and extraction protocol. Input data typically consists of a multi-dimensional vector representing time-discretized measurements. This vector is converted into a piece-wise constant function $I(t)$, with constant step length $T$ (a sketch of this is shown in Fig.~\ref{Fig_masking_sketch} (a)). While in principle, the reservoir can be directly driven with the piece-wise constant input data $I(t)$, this usually leads to a comparably low-dimensional trajectory for delay-systems. Efficient reservoir computing with delay-systems relies on the so called 'masking' or 'time-multiplexing' procedure \cite{APP11}. A $T$-periodic function, called the mask, is multiplied on top of the piece-wise constant $I (t)$ resulting in a rich input $\lambda (t)$, shown in Fig.~\ref{Fig_masking_sketch} (b). This more complex input data stream induces a dynamically richer response of the reservoir that is still strongly dependent on the input data. 

Using a single-node with delay as a reservoir therefore has a few distinct advantages: The setup is easily scaled up or down, depending on the required phase space dimension if the delay line is simply modified. Furthermore, these systems have been successfully used in experimental setups owing to the comparably simple implementation when only a single 'active node' is needed \cite{APP11, BRU13a, LAR12}. However, the sequential nature of the data input and readout also slows the system down. In fact, doubling the number of virtual nodes would lead to a halfing of the clock cycle. 

\section{Virtual and Mixed Networks}
\label{SEC_virtual_and_mixed_networks}

Often the mask is chosen to be a piece-wise constant function with lengths of $\theta = T/N_{V}$, where $N_V \in \mathbb{N}$ is the time-multiplexing or virtualization factor. However, this is not the only choice \cite{NAK16}. As with the input, the readout also needs a reference clock when used in a continuous time system. This neccessarily needs to align with the input periods $T$, as otherwise input and output would start to drift with respect to each other. The output timings could now be done once per input-timing window $T$, but this would lead to a very poorly resolved and low-dimensional readout of the complex phase space trajectory. With a piece-wise constant mask it is much more natural to read out with the same frequency as the characteristic mask time scale, i.e. once per $\theta$. Reading out even faster is possible, however in real experiments this readout process is the actual bottle neck and therefore increasing it is not trivial. With a piece-wise constant mask and synchronized read-out the system can essentially be thought of as a 'Virtual Network'. Each time interval $\theta$ represents a 'Virtual Node' of which there are $N_V$ in total \cite{APP11}. This analogy helps link the original network-based concepts of reservoir computing with the delay-based examples. For some cases, an explicit or approximate transformation to a network picture can be derived and used \cite{GRI15, SCH13l}. 

\begin{figure}
  \begin{center}
    \includegraphics[width=0.47\textwidth]{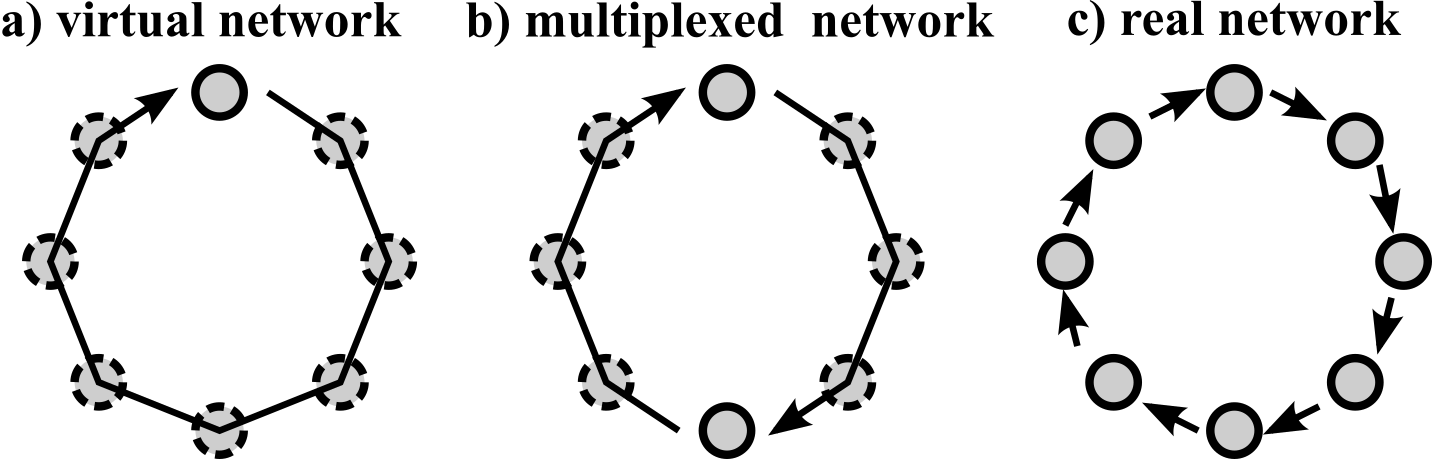}
  \end{center}
\caption{Visualization for the creation of a) 'Virtual Networks' in systems with delay, b) A mixture of both Virtual and Real network elements and c) a network of pure real oscillators.}
\label{Fig_real_vs_virtual_networks}
\end{figure}

In this work we systematically compare network and delay-based approaches. For this, we construct what we call 'multiplexed networks' that include both delay lines and real nodes within a networks. We refer to physical nodes as 'real nodes', as opposed to the 'virtual nodes' created through time-multiplexing. With some small adjustments the masking procedure described in Sec.~\ref{SEC_rc_with_del} can be generalized to coupled network motifs.  In principle one could take any small network of instantaneously coupled oscillators, give each individual node or a subset of nodes its own delayed self-feedback with identical delay time $\tau$ and then apply the masking procedure to the network as a whole. This way, the network could be seen as a single multidimensional node that is used in the same way as described in Sec.~\ref{SEC_rc_with_del}. However, this approach has two drawbacks: First, the network motif needs additional external feedback connection, which would neccessitate additional physical components for an experimental implementation. Additionally, most sufficiently fast real-world implementations of coupled systems will not be able to be instantaneously coupled inside the network motif. We propose a different masking process for network motifs profits that profits from the fact that small network motifs of real nodes would usually already contain time-delay connections. Followingly these already present delay-connections can be utilized for the time-multiplexing procedure. We therefore generate a mask for each node independently and simply drive and read-out the network as is. While in the traditional delay-line approach the 'virtual nodes' are often portrayed as lieing on the single long delay-loop, this representation is no longer possible in a complex network motif. Nevertheless, the same principles of time-multiplexing apply.  

Figure~\ref{Fig_real_vs_virtual_networks} shows a sketch of the different network types that we will compare: Virtual networks consist of a single node with feedback, in which a rich phase space response is created with the masking procedure as described in Sec.~\ref{SEC_rc_with_del}, cf. Fig.~\ref{Fig_real_vs_virtual_networks}~a). We also look at delay-coupled network motifs using the already present delay of the connections, shown in Fig.~\ref{Fig_real_vs_virtual_networks}~b). A mask of length $T$ is generated for each node individually, and the state of every real node is recorded simultaneously. When we increase the number of real nodes and reduce the virtualization factor $N_V$ to 1 the system becomes a network of purely real oscillators, Fig.~\ref{Fig_real_vs_virtual_networks}~c).

Many systems in nature are coupled oscillatory systems. These not only include electromagnetic waves, but also nanomechanical oscillators and chemical oscillators among others. As we are interested in fundamental properties of reservoir computing systems, we will not focus on a specific experimental application in depth. Instead, we employ the fundamental case of $N_R$ delay-coupled Stuart-Landau oscillators, described by the complex variables $Z_{k} \in \mathbb{C}$ in the following system of DDEs: 
\begin{align}        
    \dot{Z}_k &=(\lambda + \ii \omega + \gamma\, |Z_k|^2) Z_k + \kappa e^{\ii \phi} \sum_{l = 0}^{N_R}  G_{kl} Z_l(t-\tau)
    \label{model_equation}
\end{align}
Here $\lambda \in \mathbb{R}$ is the bifurcation parameter with an Andronov-Hopf-bifurcation occurring at $\lambda = 0$ in a solitary oscillator, $\omega \in \mathbb{R}$ is the frequency of the free-running oscillator. 
The sign of the real part of the nonlinearity $\gamma \in \mathbb{C}$ defines whether the Andronov-Hopf-bifurcation is sub- or supercritical, while the imaginary part defines the hardness of the spring and induces an amplitude-phase coupling. 
Hence, $\textbf{Im}(\gamma)$ is linked to the amplitude-phase or linewidth-enhancement factor of semiconductor lasers \cite{BOE15}. 
The network-coupling between the oscillators is defined by the coupling strength $\kappa \in \mathbb{R}$ and coupling phase $\phi \in [0, 2\pi]$. The topology of the network is given by the adjacency matrix $G_{kl}$. The coupling and feedback terms $Z_k (t - \tau)$ are delayed with the delay time $\tau$. For our numerical simulations we set $\mathbf{Re}(\gamma) = -0.1$ (supercritical case), $\mathbf{Im}(\gamma) = 0.5$, $\omega = 1$  and $\kappa = 0.1$, unless noted otherwise. We assume all delay-lengths to be identical.
This model can approximate a wide range of different oscillatory systems that are coupled instantaneously, i.e., with negligible transmission and coupling delay. The Stuart-Landau system is the normal form of an Andronov-Hopf-bifurcation and therefore any system close to such a bifurcation can be approximated with the nonlinearity of Eq.~\eqref{model_equation}. This model can therefore also describe lasers if they are operated close to an instability threshold. 

For the systematic study of reservoir computing performance, we create networks of different sizes. As the reservoir computing performances generally increases with the dimension of the read-out, we keep the output-degrees constant. For this, we create networks for which the product of real nodes $N_R$ and degree of virtualization $N_V$ is constant. However, as the degree of virtualization is linked with the time-per-virtual-node $\theta$, this means we also change the delay time $\tau$ when changing $N_V$. We have chosen to use a system with base $2$ to create the different networks. We keep the product $N_V \, N_R = 2^8$ constant and increase $N_R$ by factors of 2 from $N_R = 1$ to $N_R = 2^8$. The time per virtual node is $\theta = 12$ and the delay time set to $\tau = 17 * N_V$. Our mask length and delay-length are therefore non-identical, which has been shown to increase performance \cite{PAQ12}. We inject by varying $\lambda$ in Eq.~\eqref{model_equation}, corresponding to a driving current in a laser. The maximum injection strength is $0.01$ and the mask values are binary, i.e. either $0$ or $1$.

We test the system using the Nonlinear-Autoregressive Moving Average Task (NARMA) \cite{ATI00} of length 10. This simulates a complex nonlinear transform of an input array, where both memory and nonlinear transformation capabilities are needed. From a given series series $u_k$ drawn from a uniform distribution $[0, 1]$, the trained system has to calculate the corresponding NARMA10 series. The NARMA10($u_k$) is defined by an iterative formula $A_k$ as given by:
\begin{align}        
   A_{k+1} &= 0.3 A_k + 0.05 A_k \left(\sum_{i = 0}^{9}  A_{k - i}\right) + 1.5 u_{k-9} u_k + 0.1
    \label{EQ_narma10}
\end{align}

Furthermore, we also test the performance for the Santa-Fe laser chaotic time series prediction task. This dataset contains roughly 9000 data points for a chaotic laser. Given the timeseries up to a point $t$, the system is trained to predict the future step(s) of this chaotic series. We restrict our-selves to the 1-step prediction in this report. We have always used a training and testing length of 2900 datapoints with a buffer length of 100 for the Santa Fe task.

We evaluate the performance for both tasks by calculating the normalized-root-mean-squared error (NRMSE), where the normalization is done with the variance $\sigma^2$ of the target series. Given a target series $y_k$ and the output of the trained system $\hat{y}_k$, we calculate the NRMSE as:
\begin{align}        
   NRMSE(y, \hat{y}) = \sqrt{\frac{\sum_{k} \left(y_k - \hat{y}_k\right)^2}{\sigma^2(y)}}.
    \label{EQ_NRMSE}
\end{align}

The NRMSE is $0$ for a perfect agreement between target and output, while $1$ is the highest reasonable error, representing a static prediction of the average of the target.

\section{Two delay-coupled Oscillator}
\label{sec_2_oscillators}

\begin{figure}
  \begin{center}
    \includegraphics[width=0.47\textwidth]{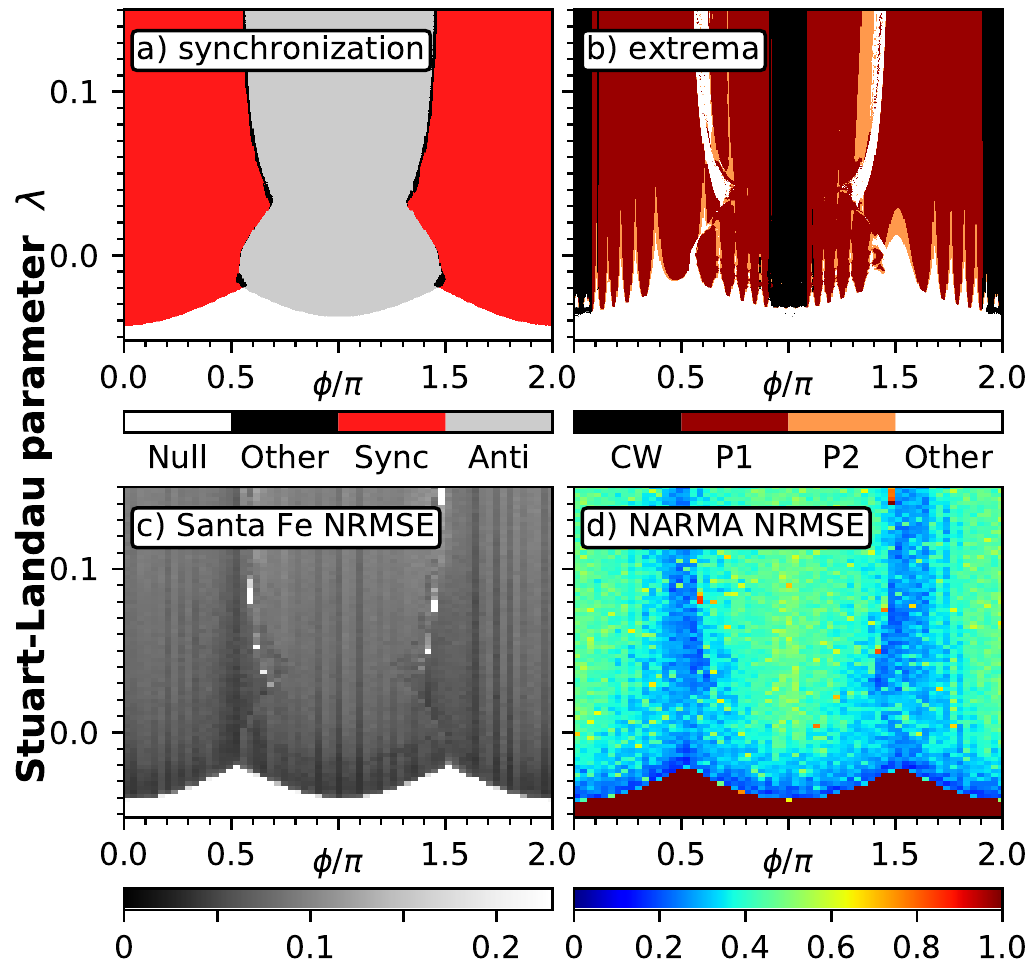}
  \end{center}
\caption{Two delay-coupled oscillators: Two dimensional parameter scan for the system as shown in Eq.~\eqref{model_equation} for different pump terms $\lambda$ and coupling phase $\phi$. (a) Synchronization state of the system; synchronized (red), antisynchronized (grey), off-solution $Z_k = 0$ (white). b) Periodicity of the dynamics: Harmonic oscillations (black), regular amplitude oscillations (red), higher order dynamics (yellow) and the off-solution (white). Reservoir computing performance measured by NRMSE in color code for the Santa Fe 1-step prediction c) and NARMA10 task d). Parameters: $N_R=2$, $N_V = 128$, $\tau = 2176$, $\gamma = - 0.1$, $\omega = 1$, $\kappa = 0.04$.}
\label{Fig_dynamics_and_performance}
\end{figure}

The simplest case of a multiplexed network is the case of two real delay-coupled nodes. The system of coupled Stuart-Landau oscillators is a well-studied example in nonlinear dynamics. Networks of such oscillators can exhibit a wide range of different dynamics  \cite{HAK92, NAK93, KU15, KOS13, USH05, GEF14, ZAK14, ZAK15b}. Here, we study the connection between reservoir computing capabilities and the dynamics of the underlying network \cite{ROE17}. In Fig.~\ref{Fig_dynamics_and_performance} we have numerically integrated the system given by Eq.~\eqref{model_equation} with $N_V = 128$, $N_R = 2$, $\tau = 2176$, $\gamma = - 0.1$, $\omega = 1$, $\kappa = 0.04$ for different values of the coupling phase $\phi$ and base input parameter $\lambda$. Note, that $\lambda$ is additionally modified by the input procedure by up to $0.01$ in panels c) and d). 

Panel a) of Fig.~\ref{Fig_dynamics_and_performance} shows the synchronization type of the network without input. White regions correspond to the off-state $Z_1 = Z_2 = 0$, synchronization $Z_1 = Z_2 \not = 0$ occurs in the red regions centered around $\phi = 0$ and anti-synchronization $Z_1 = - Z_2 \not = 0$ for the grey regions. Fig.~\ref{Fig_dynamics_and_performance}~b) shows the number of different maxima of $|Z(t)|$ of the network without input, highlighting the regions of dynamic complexity. The black regions exhibit constant amplitudes $|Z_N| = c$,  while the colored regions contain higher-order dynamics, i.e. amplitude oscillations, period doubling cascades and quasiperiodic behaviour. The white regions contain other dynamics, mostly the off-solution and complex behaviour. Finally, Fig.~\ref{Fig_dynamics_and_performance}~c) shows the error landscape as measured by the NRMSE for the Santa Fe 1-step prediction task for this network, with the results for the NARMA10 task in panel d). We have used a training length of 1500 data points  for training and 500 data points for testing, with an additional 150 data points as a buffer for the NARMA task. The darker/blue colors correspond to a low error, i.e. high performance, while brighter/yellow regions exhibit poor reservoir computing capabilities. 

Analyzing the relationship between the different characteristics shown in Fig.~\ref{Fig_dynamics_and_performance} we can find a few general trends: First, we find that the regions of the off-solution (low $\lambda$, compare Fig.~\ref{Fig_dynamics_and_performance}~a) cannot be used for reservoir computing. This is not surprising, as the system does not react at all to input, if the parameter $\lambda$ never exceeds the onset of oscillations. A system that does not react to the driving signal will not be able to output a transformation of that signal and hence not have any computational power. Next, within the regions of synchronization and desynchronization there is considerable variability of the error in Fig.~\ref{Fig_dynamics_and_performance} c) and d). The regions of lowest error generally lie within the area of synchronization, while the anti-synchronization never reaches NRMSE values that low. Moreover, there exist many regions that exhibit time-dependent amplitude-modulations even without input (cf. colored regions in Fig.~\ref{Fig_dynamics_and_performance}~b), i.e. the system is on a limit cycle. These amplitude oscillations will, in general, not have a period that is identical to our input timing window $T$. Hence, the network will react differently to the same input, depending on its position in phase space when the input is applied. This violates one of the core requirements for reservoir computing, namely the 'reproducibility' of phase-space trajectories\cite{OLI16}. Nevertheless, we find the regions of best performance to lie in those areas. This demonstrates that looking at the network without input is not sufficient to predict the reservoir computing behaviour. Furthermore, it is likely that the amplitudes of the oscillations can influence, how much the performance is degraded. Fig.~\ref{Fig_dynamics_and_performance}~c) and~d) furthermore reveal, that the regions of best performance can be found close to the bifurcation lines separating the regions of different behaviour. The 'edge of chaos' has been mentioned as the optimal driving point for reservoir computers in previous works \cite{BER04, VER07}. A similar effect is occuring here, where the regions closer to dynamic complexity exhibit a better performance.

\section{Networks}

\begin{figure}
  \begin{center}
    \includegraphics[width=0.47\textwidth]{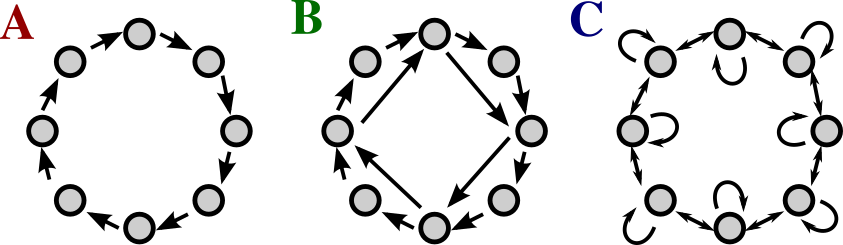}
  \end{center}
\caption{Sketch of the network types used for the multiplexed networks. A: Unidrectional ring. B: Unidrectional Ring with forward links from every fourth element four units ahead. C: Bidirectional ring with self-feedback.}
\label{FIG_network_types}
\end{figure}

When going beyond the simple case of $N_R =2$, the possibility of how to couple the networks increases dramatically. We cannot test and investigate all topologies, so in this report we use three different topologies for the underlying network of real oscillators. As we are using the already present links inside the network, we can in principle use any network topology in conjunction with the time-multiplexing procedure. However, as we are most interested in general trends we will abstain from using random or complex topologies and focus on three very simple topologies as sketched in Fig.~\ref{FIG_network_types}. First, we test a unidirectional ring. This is not only a very simple network topology, but also in some sense represents the 'virtual network' created for a single node with delay \cite{SCH13l}. An example is shown in Fig.~\ref{FIG_network_types}~A for $N_R = 8$. As a second example, we use the same network but add crosslinks for every fourth node, jumping forward 4 nodes. Fig.~\ref{FIG_network_types}~B shows an example for $N_R = 8$ (the sketch contains jumps with only length 2). Note, that this network will be identical by construction to the unidirectional ring for $N_R \leq 4$. For both types of unidirectional ring we take the links to be all identical in strength. Lastly, we test a bidirectional ring with self-feedback, as shown in Fig.~\ref{FIG_network_types}~C. Here we are inspired by the often used difference coupling $Z_{i+1}(t - \tau) - Z (t -\tau ) $ and hence the self-feedback is assumed to have double the strength and a differing sign than the bidirectional links.

For each parameter combination we randomly generate the mask sequence, i.e. the two-dimensional scans shown have a different mask for every parameter combination. The NARMA sequence is fixed. In the following we have used a training length of 5000, with an additonal buffer of 1000 at the start to let the system settle into the correct trajectory. The evaluation was done with identical lengths. The simulations were programmed with custom code written in C++ and run on the CPUs of a network of approximately 30 conventional workstation computers.

\begin{figure}
  \begin{center}
    \includegraphics[width=0.47\textwidth]{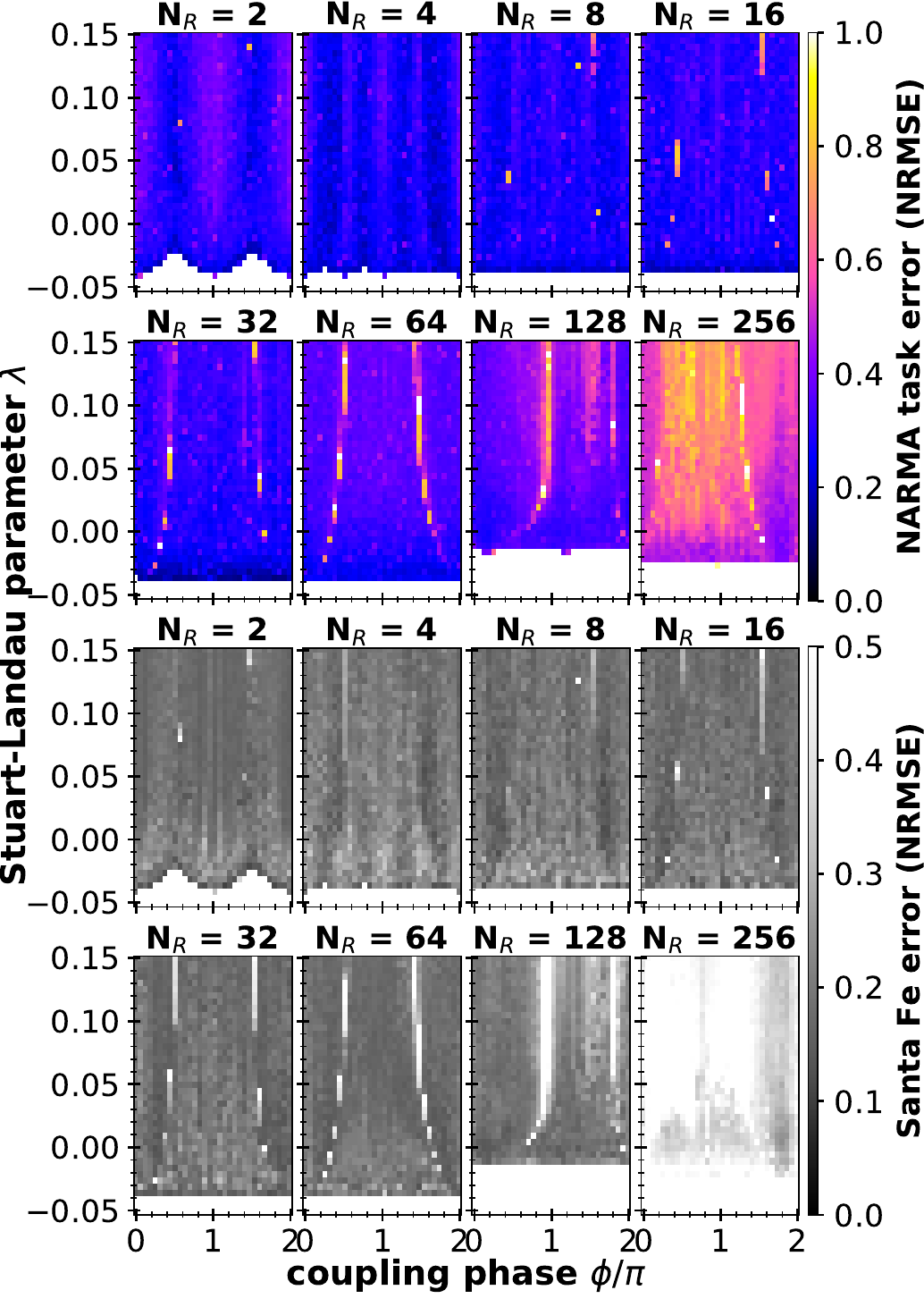}
  \end{center}
\caption{Unidirectional Ring: Numerically evaluated reservoir computing capabilites of the NARMA task (top) and Santa Fe task (bottom) for different $N_R$ plotted in the plane of $\lambda$ and coupling phase $\phi$. The color code shows the root mean squared error (NRMSE) as described in Eq.~\eqref{EQ_NRMSE} for the NARMA task in blue-yellow, and the Santa Fe task in grey. Parameters: $N_V = 256/N_R$, $\tau = 17*N_V$, $\gamma = - 0.1$, $\omega = 1$, $\kappa = 0.04$.}
\label{FIG_uni_ring_RC}
\end{figure}

Figure~\ref{FIG_uni_ring_RC} shows the results of our numerical simulation of networks with different number of real nodes $N_R$  for the unidirectional ring as shown in Fig.~\ref{FIG_network_types}~A, while adjusting the number of virtual nodes to keep the overall reservoir readout dimension constant. The top 8 panels show the result for the NARMA10 task, while the bottom panels show the results for the Santa Fe laser 1-step prediction task. High errors (yellow/white) designate undesirable regions for computing, while low errors (blue/dark grey) show regions of effective computing. 

The structure of the colored regions in Fig.~\ref{FIG_uni_ring_RC} is dominated by a sudden cutoff for low $\lambda$. This corresponds to the threshold of oscillations for the individual oscillator, and for values lower than a critical $\lambda$ no stable oscillation exists. This is the same mechanism as already discussed for the case of $N_R = 2$ in Sec.~\ref{sec_2_oscillators}. Note, that the maximum input strength of the pump term is $0.01$. The comparison of the different number of real nodes in Figure~\ref{FIG_uni_ring_RC} shows several trends: While at first a clear global structure is visible for $N_R = 2$ with two 'valleys' of good performance, this splits into four valleys for $N_R = 4$. For larger networks this structure is washed out, and instead two lines of high NRMSE become visible (the red and yellow lines in Fig.~\ref{FIG_uni_ring_RC} for $N_R = 64$). These likely indicate bifurcations in the underlying state diagram. For the networks with the highest number of real nodes ($N_R = 128$ and $N_R = 256$ in Fig.~\ref{FIG_uni_ring_RC}) the performance is greatly degraded. Additionally, the cut-off threshold is strongly modified in these last panels, as the delay-time $\tau$ is significantly shorter, as we scale the delay inversely with the number of real nodes. 

Overall, the performance of the network reaches values favorable for reservoir computing in all the low-$N_R$ cases. Furthermore, from $N_R = 16, 8$ in Fig.~\ref{FIG_uni_ring_RC} it is qualitatively apparent, that the dependence of the performance on the parameters is reduced for these intermediate networks.


The lower 8 panels of Fig.~\ref{FIG_uni_ring_RC} show the results for numerical simulations of the network of Stuart-Landau oscillators as given by Eq.~\eqref{model_equation} for the Santa Fe laser chaotic time-series prediction task for different $N_R$ in the plane of $\lambda$ and coupling-phase $\phi$. The grey color code shows the test run NRMSE. Note the different scaling compared to the NARMA10 color scale, due to the overall lower NRMSE for the Santa Fe task. The Santa Fe laser task NRMSE shows a qualitatively similar behaviour as the NARMA10 NRMSE shown in the top-panels of  Fig.~\ref{FIG_uni_ring_RC}. The same lines of high error can be found  for $N_R = 32, 64, 128$. Additionally, the region of good performance is limited by the same cut-off for low $\lambda$.  We find a high degree of similarity between the performance in the Santa Fe and NARMA10 task, i.e. regions that are suitable for one are also suitable for the other. For some values of $N_R$ this is more apparent than others in Fig.~\ref{FIG_uni_ring_RC}. The generally accepted mechanism behind most of the transformations is the ability of the reservoir to store memory and its ability of nonlinear transformation. Ref.~\cite{DAM12} introduced the notion of dividing the stored information inside the reservoir into the linear memory capacity, representing a mere recording of past inputs, and nonlinear memory capacity, representing storage of transformed information. Together these memory capacities form a complete basis in the space of transformations. The fact that both Santa Fe and NARMA10 exhibit similar performance profile as shown in Fig.~\ref{FIG_uni_ring_RC} indicates that both tasks need a similar set of linearly and nonlinearly stored information. 

We have used the data sets generated for Fig.~\ref{FIG_uni_ring_RC} to calculate the covariance of NRMSE for the Santa Fe and NARMA10 task. Using the raw data, we find very high covariances of over $0.9$ for all values of $N_R$. However, this behaviour is mostly dominated by the regions of NRMSE close or equal to $1$ for low $\lambda$. When we exclude all points with NRMSE greater than $0.9$ we get drastically lower values, ranging from $0.15$ for $N_R = 4$ to $0.75$ for $N_R = 128$. For $N_R = 2$ we find a small negative covariance of $-0.13$. These much lower values have two reasons: First, even a visual inspection does reveal some deviation of details for the NRMSE landscape in Fig.~\ref{FIG_uni_ring_RC}. For example, the error seems to generally increase with $\lambda$ for $N_R  = 64$ for the NARMA10 task, while it decreases for the same $N_R$ with $\lambda$ in the Santa Fe task. Second, and more importantly, the covariances we have calcuated only represented a lower bound. We have two sources of uncertainty for our simulations that we cannot control for in Fig.~\ref{FIG_uni_ring_RC}, namely that we are independently creating random binary masks and drawing the source sequence $u_k$ for the NARMA10 task for every parameter combination and $N_R$. We are therefore comparing the Santa Fe and Narma task results for different masks, while the parameter scan of the NARMA10 task shown in Fig.~\ref{FIG_uni_ring_RC} uses a new independet but identically distributed $u_k$ for every parameter combination. These limitations have only been considered after the extensive numerical simulations required for the two-dimensional parameter scans and therefore a detailed analysis will have to be left for future investigations.

\begin{figure}
  \begin{center}
    \includegraphics[width=0.47\textwidth]{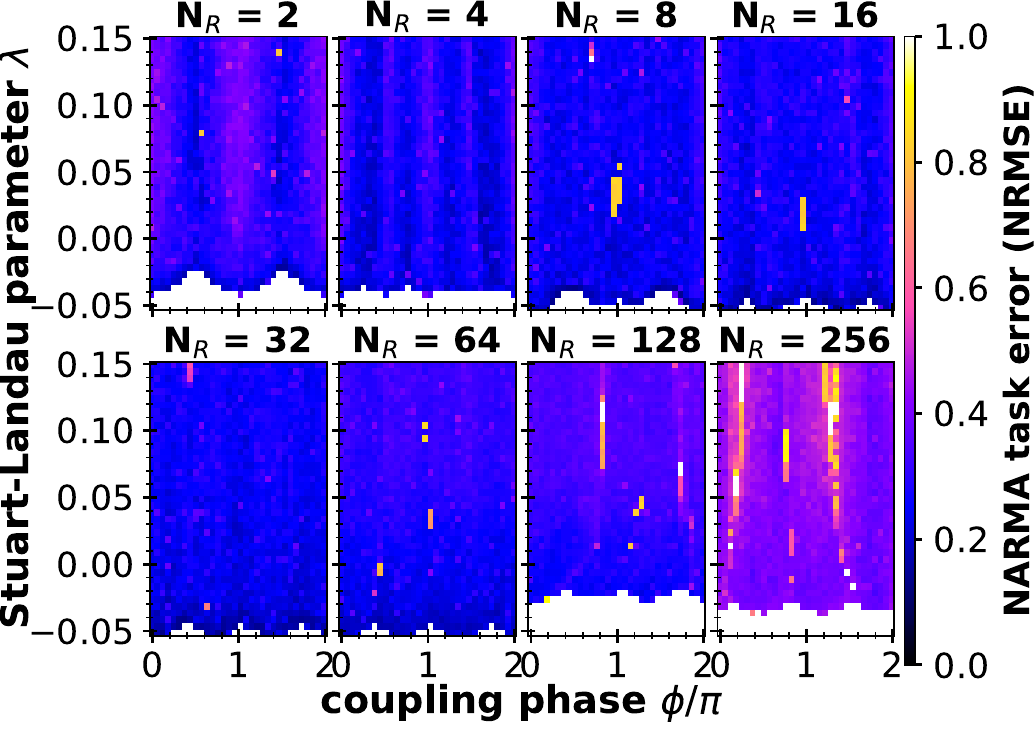}
  \end{center}
\caption{Unidirectional Ring with Jumps: Numerically evaluated reservoir computing capabilites of the NARMA task for different $N_R$ plotted in the plane of $\lambda$ and coupling phase $\phi$. The color code shows the root mean squared error (NRMSE) of the NARMA10-task as described in Eq.~\eqref{EQ_NRMSE}. Parameters: $N_V = 256/N_R$, $\tau = 17*N_V$, $\gamma = - 0.1$, $\omega = 1$, $\kappa = 0.04$.}
\label{FIG_uni_with_jumps_ring_narma}
\end{figure}

Figure~\ref{FIG_uni_with_jumps_ring_narma} shows the result of the NARMA testing error for the unidirectional ring with jumps (network type in Fig.~\ref{FIG_network_types}~B). Due to construction, this system is identical for $N_R = 2$ and $N_R = 4$ to the pure unidirectional ring as shown in Fig.~\ref{FIG_uni_ring_RC}. We do however randomly generate a new mask and NARMA target series for every parameter combination. Therefore a comparison between Fig.~\ref{FIG_uni_with_jumps_ring_narma} and Fig.~\ref{FIG_uni_ring_RC} for $N_R = 2$ and $N_R = 4$ allows us to see the influence of differing masks. The global structure of the NRMSE is not changed. This indicates that the performance is reproducable across different masks and NARMA10 series.  For $N_R \geq 8$ Fig.~\ref{FIG_uni_with_jumps_ring_narma} and Fig.~\ref{FIG_uni_ring_RC} differ due to the extra links added in Fig.~\ref{FIG_uni_with_jumps_ring_narma}. From a mere visual inspection no drastic difference of the quality of perfomance can be seen. Nevertheless, the global structure differs as the lines of bad performance and border regions with the off-state have shifted. This is to be expected, as additional links will change the bifurcations occuring in the network and bifurcations are usually associated with extrema in the performance. As was found for the pure unidrectional ring, we also see a dramatic breakdown in performance in Fig.\ref{FIG_uni_with_jumps_ring_narma} for $N_R = 256$.  Additionally, the same 'washing out' of structure can be observed for the intermediate values of $N_R = 8, 16, 32$, indicating a reduced parameter dependence of multiplexed networks with multiple real and virtual nodes.

\begin{figure}
  \begin{center}
    \includegraphics[width=0.47\textwidth]{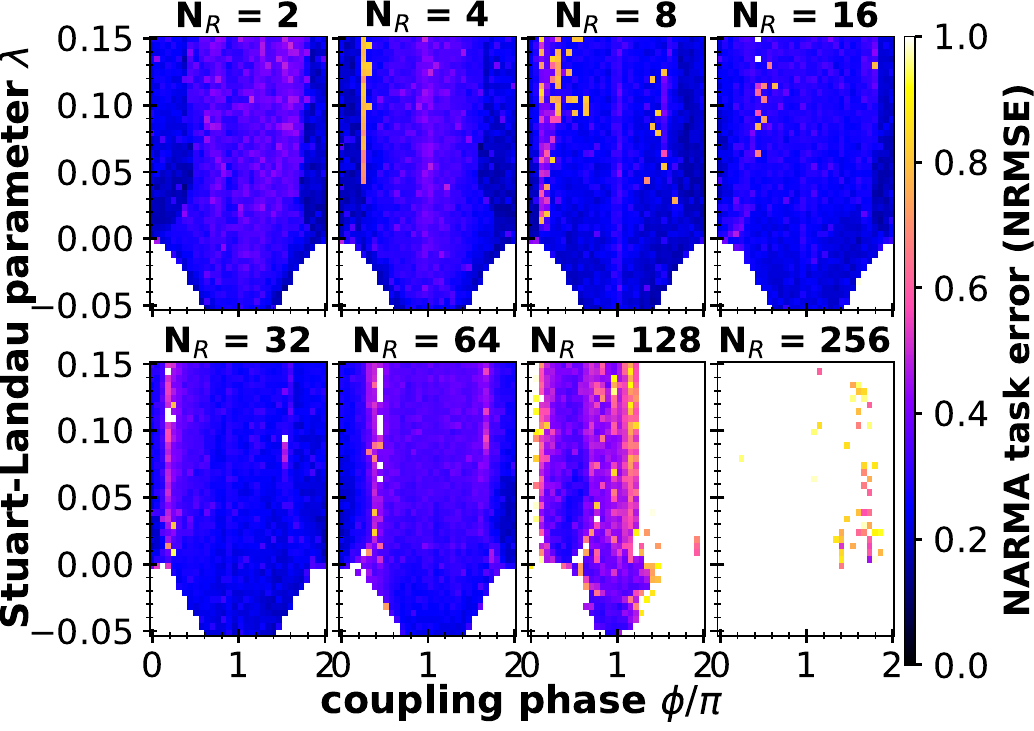}
  \end{center}
\caption{Bidirectional Ring with Self-Feedback: Numerically evaluated reservoir computing capabilites of the NARMA task for different $N_R$ plotted in the plane of $\lambda$ and coupling phase $\phi$. The color code shows the root mean squared error (NRMSE) of the NARMA10-task as described in Eq.~\eqref{EQ_NRMSE}. Parameters: $N_V = 256/N_R$, $\tau = 17*N_V$, $\gamma = - 0.1$, $\omega = 1$, $\kappa = 0.04$.}
\label{FIG_bid_ring_narma}
\end{figure}

Lastly, Fig.~\ref{FIG_bid_ring_narma} shows the performance of a bidirectionally coupled ring of oscillators with self-feedback as described by Eq.~\eqref{model_equation} (topology as sketched in Fig.~\ref{FIG_network_types}~C). Due to the fundamentally different topology, the global structure of the NRMSE shown in Fig.~\ref{FIG_bid_ring_narma} differs from Fig.~\ref{FIG_uni_ring_RC} and Fig.~\ref{FIG_uni_with_jumps_ring_narma}. Both the boundary towards the 'off-solution' (white areas of poor performance at the bottom of Fig.~\ref{FIG_bid_ring_narma}) as well as regions of optimal performance are at different locations. The dropout of performance for low degrees of virtualization, i.e. high number of real nodes is the most extreme in this topology (compare $N_R = 256$ across Fig.~\ref{FIG_uni_ring_RC}, \ref{FIG_uni_with_jumps_ring_narma} and~\ref{FIG_bid_ring_narma}). This is possibly due to higher multistability of the system due to the more complex but regular topology.

\begin{figure}
  \begin{center}
    \includegraphics[width=0.47\textwidth]{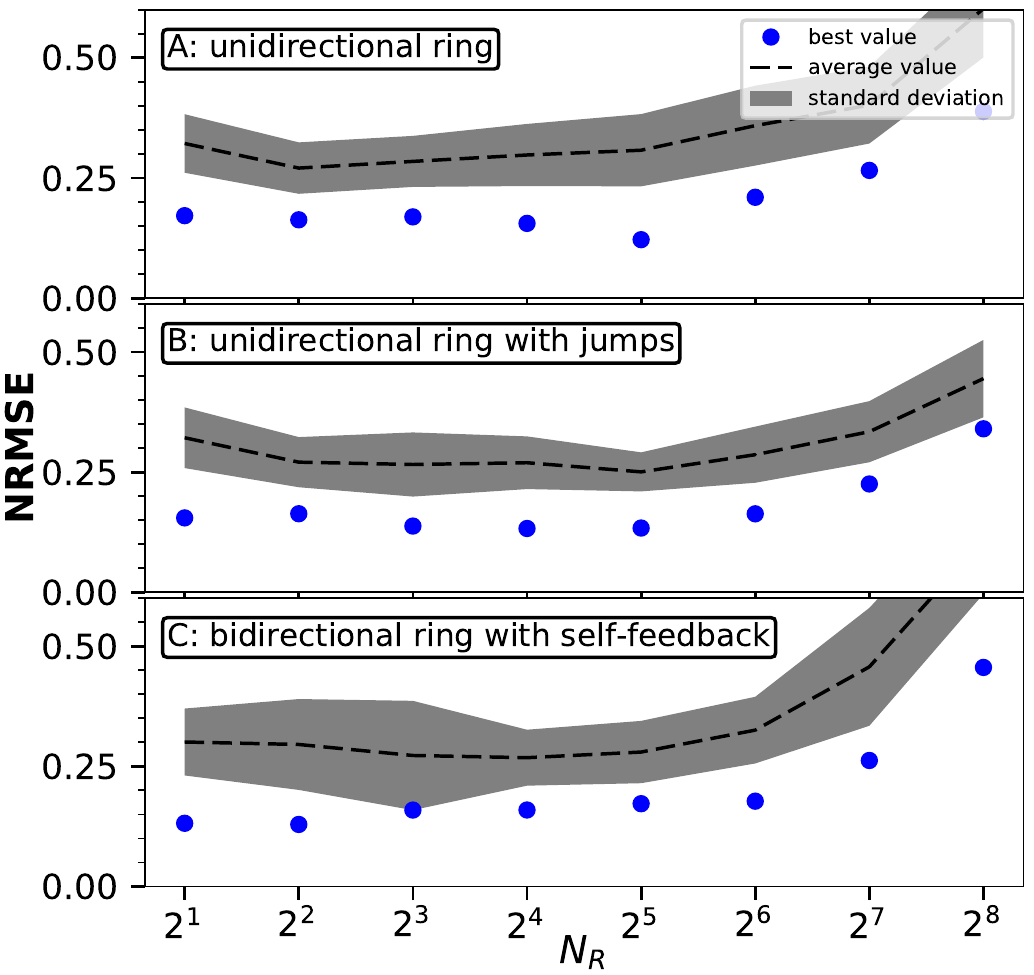}
  \end{center}
\caption{NARMA performance measured by the NRMSE as a function of the number of real nodes $N_R$ for the unidirectional ring (A), unidirectional ring with jumps (B) and the bidirectional ring with self-feedback (C). The best value found is shown in blue, while the average is shown by the black dashed line. The grey area represents one standard deviation from the average. }
\label{FIG_V_vs_NRMSE}
\end{figure}

For a qualitative comparison of the general trends we have generated Fig.~\ref{FIG_V_vs_NRMSE}, where the number of real nodes and NRMSE for the NARMA10 task is shown. We have used data simulated for the 2-parameter scans for the unidirectional ring shown in Fig.~\ref{FIG_uni_ring_RC}, \ref{FIG_uni_with_jumps_ring_narma} and~\ref{FIG_bid_ring_narma}. The blue dots correspond to the optimal or lowest NRMSE found within the 2d-scan for a given network size $N_R$. For the simulations as obtained Fig.~\ref{FIG_uni_ring_RC} we have found that the network with no virtualization never showed a NRMSE smaller than $0.4$. The black line in Fig.~\ref{FIG_V_vs_NRMSE} shows the average NRMSE for the NARMA task (excluding the white regions of no performance in Fig~\ref{FIG_uni_ring_RC}). The grey band shows one standard deviation of the NRMSE data. The qualitative impression of the color plots in Fig.~\ref{FIG_uni_ring_RC} are largely validated by the quantitative evaluation: Both the average NRMSE as well as the optimal NRMSE are mostly flat for multiplexed networks created for low number of real nodes $N_R$ until the performance breaks down for the large networks with low virtualization. All network topologies show qualitatively similar results in Fig.~\ref{FIG_V_vs_NRMSE}.


There are still more effects that warrant some attention. We run a few simulations for a network of slightly non-identical units for the bidirectional ring with self-feedback (Fig.~\ref{FIG_network_types}~C). This more closely resembles an actual experimental implementation of a network, as in real setups no two nodes would be absolutely identical. We have used the same parameter locations as for the bidirectional ring with self-feedback. However, the difference we found was small. 

For all the networks used here, we have employed the same delay length for every network link. Considering that making the mask and delay term non-identical improves performance, it seems likely that non-identical delay-lines could have a similar effect. However, simulating a network of non-identical delay-links is time-consuming and left for future investigations here. 

\section{Conclusions}
\label{sec_Conclusion}

We have investigated the reservoir computing performance of a time-continuous system with delay. While many studies have been published concerning a single dynamical system with a long delay loop, we have numerically simulated network motifs consisting of several nodes that are delay-coupled. We have used the time-multiplexing/masking procedure to generate additional high-dimensional trajectories. We have constructued the 'multiplexed networks' in such a way, that the over all dimension of the read-out stays constant. This enables us to not only qualitatively, but also quantitatively compare the reservoir computing performance.

Reservoir computers consisting exclusively of large real, regular networks have exhibited poor performance in both the NARMA and Santa Fe task, independent of local topology. We attribute this both to the higher multistability of such systems, as well as the relative lack of complex phase space trajectories due to the absence of time-multiplexing. In contrast, networks of only small and intermediate size have performed consistently on a state-of-the-art level. We found a lowered parameter sensitivity and an enhanced computation speed for such systems. This is encouraging for experimental realizations, as our results indicate that the design of a reservoir computer can be chosen with some degree of freedom. As long as a sufficient time-multiplexing is used, the number of real nodes can be adjusted to fit experimental limitations and desired output speed.

\section{Acknowledgement}

The authors thank B. Lingnau and L. Jaurigue for fruitful discussions. This work was supported by the DFG in the framework of the SFB910.

\section*{References}

\section{Methods}

A minimal example of an 'echo state network', i.e. a time-discrete reservoir computer, as described from Ref.~\cite{JAE01} is:
\begin{align}        
X (t + 1) = & f \left( W_{res} X(t) + W_{in} I(t) \right),  \label{RC_simple_example_1} \\
O (t) = & W_{out} X(t), 
\label{RC_simple_example_2} 
\end{align}
Where $X(t)$ is the state-vector of the network of maps, $W_{res}$ is the adjacency matrix of the network, $W_{in}$ is the matrix of input-coupling for the data stream $I(t)$. The time is taken to be discrete $t \in \mathbb{N}$ and the evolution of the network is given by Eq.~\ref{RC_simple_example_1}. $f(X)$ represents a local sigmoidal function that acts element-wise on $X$. The output $O(t)$ is calculated from $X(t)$ with the outcoupling weight matrix $W_{out}$. The matrices $W_{res}$, $W_{in}$ ande $W_{out}$ now have to be chosen in such a way that $O(t) = Y(X(I(t)))$ corresponds to the desired $ I \rightarrow O$ transformation. The distinguishing feature now lies in the optimization process. While conventional deep convolutional neural network learning schmes heavily focus on the training of the $W_{res}$, this matrix is assumed to be fixed for reservoir computing. In fact, training is only applied to the output weight matrix $W_{out}$, for which a simple linear regression is enough to find the optimal values \cite{JAE01}.

\end{document}